\documentclass[11pt,letterpaper]{article}

\usepackage{hyperref}
\usepackage{float}
\usepackage{pslatex}
\usepackage{apacite}
\usepackage{amsmath,amssymb}
\usepackage{graphicx}
\usepackage{subcaption}
\usepackage{caption}
\usepackage{color}
\usepackage{url}
\usepackage{todonotes}
\usepackage{mathtools}
\usepackage{stmaryrd}
\usepackage{booktabs}
\usepackage{array}
\usepackage{nicefrac}
\usepackage{url}
\usepackage{authblk}
\usepackage{caption,setspace}
\usepackage{xcolor}
\usepackage{changepage}
\usepackage{subcaption}
\usepackage{rotating}
\usepackage{color}
\graphicspath{{./figures/}}

\definecolor{Red}{RGB}{178,34,34}
\definecolor{Green}{RGB}{10,200,100}
\definecolor{Blue}{RGB}{10,50,150}

\usepackage[letterpaper, margin=0.9in]{geometry}

\title{Visual resemblance and communicative context constrain the emergence of graphical conventions}
\date{}

\author[a,d]{Robert D. Hawkins}
\author[a]{Megumi Sano}
\author[a,b]{Noah D. Goodman}
\author[a,c]{Judith E. Fan}

\affil[a]{Department of Psychology, Stanford University}
\affil[b]{Department of Computer Science, Stanford University}
\affil[c]{Department of Psychology, University of California, San Diego}
\affil[d]{Department of Psychology, Princeton University}

\begin{document}
\maketitle

\begin{abstract}
From photorealistic sketches to schematic diagrams, drawing provides a versatile medium for communicating about the visual world. How do images spanning such a broad range of appearances reliably convey meaning? Do viewers understand drawings based solely on their ability to resemble the entities they refer to (i.e., as images), or do they understand drawings based on shared but arbitrary associations with these entities (i.e., as symbols)? In this paper, we provide evidence for a cognitive account of pictorial meaning in which both visual and social information is integrated to support effective visual communication. To evaluate this account, we used a communication task where pairs of participants used drawings to repeatedly communicate the identity of a target object among multiple distractor objects. We manipulated social cues across three experiments and a full internal replication, finding pairs of participants develop referent-specific and interaction-specific strategies for communicating more efficiently over time, going beyond what could be explained by either task practice or a pure resemblance-based account alone. Using a combination of model-based image analyses and crowdsourced sketch annotations, we further determined that drawings did not drift toward ``arbitrariness,'' as predicted by a pure convention-based account, but systematically preserved those visual features that were most distinctive of the target object. Taken together, these findings advance theories of pictorial meaning and have implications for how successful graphical conventions emerge via complex interactions between visual perception, communicative experience, and social context.



\end{abstract}

\section{Introduction}

Human communication goes well beyond the exchange of words.
Throughout human history, people have devised a variety of alternative technologies to externalize and share their ideas in a more durable, visual form. 
Perhaps the most basic and versatile of these technologies is drawing, which predates the invention of writing \cite{clottes2008cave,tylen2020evolution} and is pervasive across many cultures \cite{gombrich1989story}.
The expressiveness of drawings has long provided inspiration for scientists investigating the mental representation of concepts in children \cite{Minsky:1972ta,KarmiloffSmith:1990ty} and clinical populations \cite{Bozeat:2003hk,chen2012clock}. 
Yet current theories of depiction fall short of explaining how humans are capable of leveraging drawings in such varied ways.  
In particular, it is not clear how drawing enables the flexible expression of meanings across different levels of visual abstraction, ranging from realistic depictions to schematic diagrams.
Do viewers understand drawings based solely on their ability to resemble the entities they refer to (i.e., as images), or do they understand drawings based on shared but arbitrary associations with these entities (i.e., as symbols)? 



On the one hand, there is strong evidence in favor of the image-based account, insofar as general-purpose visual processing mechanisms are sufficient to explain how people are able to understand what drawings mean.
Recent work has shown that features learned by deep convolutional neural network models (DCNNs) trained only to recognize objects in photos, but have never seen a line drawing, nevertheless succeed in recognizing simple drawings \cite{fan2018common}. 
These results provide support for the notion that perceiving the correspondence between drawings and real-world objects can arise from the same general-purpose neural architecture evolved to handle natural visual inputs \cite{sayim2011line,gibson1979ecological}, rather than relying on any special mechanisms dedicated to handling drawn images. 
Further, visually evoked representations of an object in human visual cortex measured with fMRI can be leveraged to decode the identity of that object during drawing production, suggesting functionally similar neural representations recruited during both object perception and drawing production \cite{fan2020relating}.
Together, these findings are convergent with evidence from comparative, developmental, and cross-cultural studies of drawing perception.
For example, higher non-human primates \cite{tanaka2007recognition}, human infants \cite{hochberg1962pictorial}, and human adults living in remote regions without pictorial art traditions and without substantial contact with Western visual media \cite{kennedy1975outline} are all able to recognize line drawings of familiar objects, even without prior experience with drawings.

On the other hand, other work has supported a symbol-based account, by pointing out the critical role that conventions play in determining how drawings denote objects \cite{goodman1976languages,miller1973cross}.
What characterizes such conventional accounts is that they rely on associative learning mechanisms that operate over socially mediated experiences, rather than pre-existing perceptual competence. 
This view is supported by developmental \cite{bloom1998intention} and computational modeling work \cite{fan2020pragmatic} that has highlighted the importance of social context for explaining how people can robustly identify the referent of even very sparse drawings. 
Moreover, several pioneering experimental studies identified a key role for real-time social feedback during visual communication in driving the increased simplification of drawings over time \cite{garrod_foundations_2007,fay2010interactive}, broadly consistent with the possibility that similar pressures shaped the emergence of modern symbol systems \cite{GalantucciGarrod11_ExperimentalSemiotics,tamariz2017experimental,fay2018create}. 
Further support for the notion that the link between pictures and their referents depends crucially on socially mediated learning comes from the substantial variation in pictorial art traditions across cultures \cite{gombrich1989story} and the existence of culturally specific strategies for encoding meaning in pictorial form \cite{hudson1960pictorial,deregowski1989real,hagen1978cultural}.

\begin{figure}
\begin{center}
\includegraphics[width=0.7\linewidth]{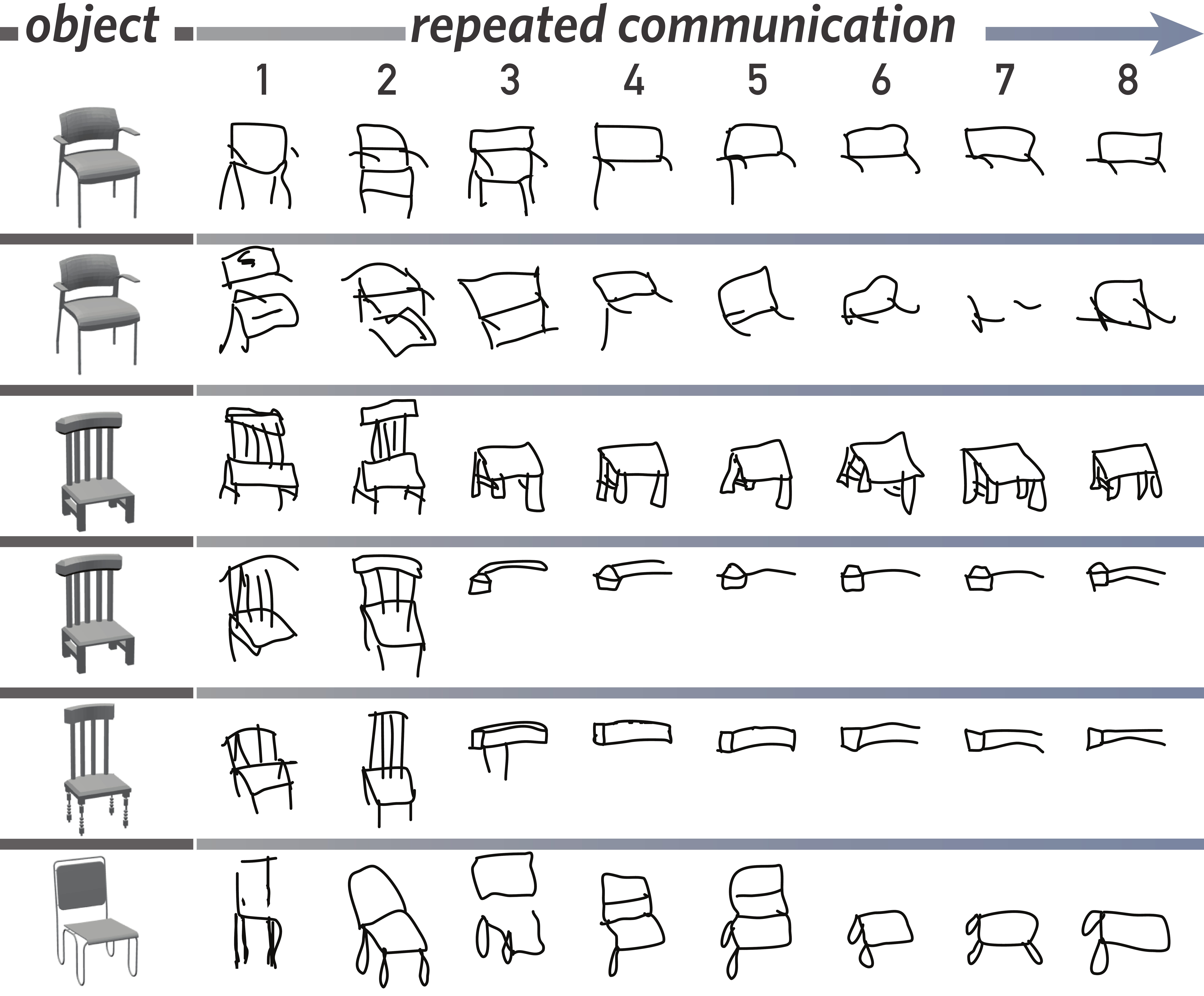}
\caption{Repeated visual communication depicting the same object.}
\label{sketch_gallery}
\end{center}
\end{figure}

In this paper, we evaluate a cognitive account of pictorial meaning that aims to reconcile these resemblance-based and convention-based perspectives.
According to this account, people integrate information from current visual experience with previously learned associations to determine the meaning of a drawing\footnote{\citeA{abell2009canny} and \citeA{voltolini2015syncretistic} have advanced related arguments in the recent philosophy literature on depiction, which continues to debate the merits of and objections to resemblance-based and convention-based views. See \citeA{kulvicki2013images} for a recent review of this debate.}. 
This account makes two key predictions:
First, while visual resemblance tends to dominate in the absence of learned associations, novel associations can emerge quickly and come to strongly determine pictorial meaning. 
For example, as two communicators learn to more strongly associate a particular drawing with an object it is intended to depict, even sparser versions of that drawing that share key visual features should still successfully evoke the original object, even if it directly resembles the object to a lesser extent. 
Second, visual resemblance will constrain the kinds of novel associations that form, such that visual information that is inherently more diagnostic of the referent will be more likely to form the basis for \textit{ad hoc} graphical conventions. 
For example, if a target object is distinguished by a particular visual attribute (e.g., a particularly long beak for a bird), then it is more likely that the sparser drawing will preserve this attribute, even at the expense of other salient attributes of the target object.




To test these predictions, we developed a drawing-based reference game where two participants repeatedly produced drawings to communicate the identity of objects in context (see Fig.~\ref{sketch_gallery}).
Our task builds on pioneering work investigating the emergence of graphical symbol systems and the importance of social feedback for establishing conventional meaning \cite{galantucci2005experimental,healey2007graphical,garrod_foundations_2007,theisen2010systematicity,garrod2010can,caldwell2012cultural,fay2010interactive,fay2013cultural,fay2014iconicity}\footnote{These drawing-based studies, in turn, belong to a broader literature studying \emph{ad hoc} convention formation in spoken language \cite{krauss1964changes,ClarkWilkesGibbs86_ReferringCollaborative}, written language \cite{hawkins2020characterizing} and gesture \cite{goldin1996silence,fay2014creating}.}, but differs substantially in focus. 
Here we are primarily concerned with understanding the cognitive constraints that enable individual sketchers and viewers to determine the meaning of pictures in context, rather than the question of where symbols come from or how symbols evolve as a consequence of cultural transmission.
As such, our tasks were designed to enable precise measurement of the visual properties of the drawings people produced, as well as the degree to which they evoked the object they were intended to depict, depending on the availability of previously learned associations. 

\begin{figure*}
\begin{center}
\includegraphics[width=1\linewidth]{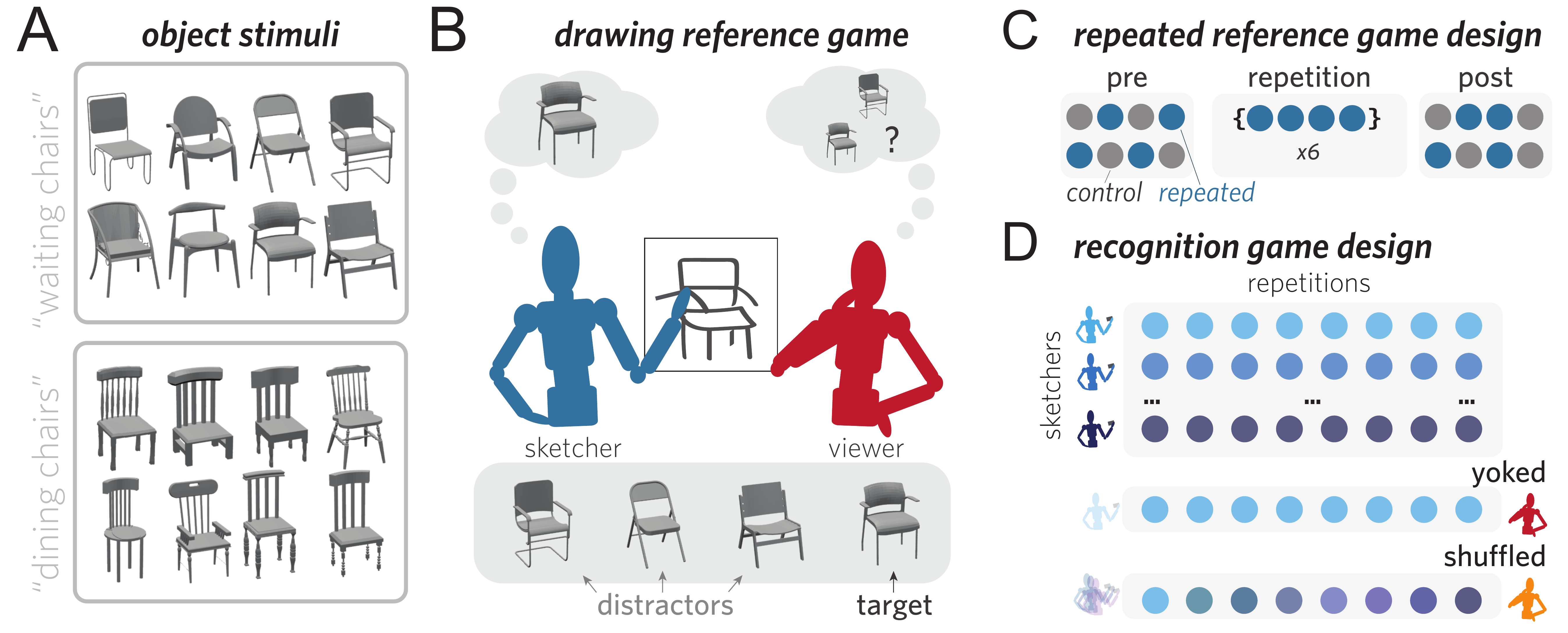}
\caption{(A) Two object collections were used, each containing eight similar objects. (B) Pairs of participants performed a drawing-based reference game in which one participant (sketcher) was cued to draw the target object such that the other participant (viewer) could identify it in context. (C) Four objects were drawn repeatedly throughout the interaction; the remaining four control objects were drawn once each at the beginning and end of each interaction. (D) Recognition participants aimed to identify the target object in context based on drawings from the reference game experiment. These drawings were either all from a single reference-game interaction (Yoked) or from all different interactions (Shuffled).}
\label{task_stimuli}
\end{center}
\end{figure*}

\section{Results}

To investigate the potential role that both visual information and shared knowledge play in determining how people communicate about visual objects, we used a drawing-based reference game paradigm.
On each trial, both participants shared a visual context, represented by an array of four objects that were sampled from a set of eight visually similar objects (Fig.~\ref{task_stimuli}A).
One of these objects was privately designated as the target for the sketcher.
The sketcher's goal was to draw the target so that the viewer could select it from the array of distractor objects as quickly and accurately as possible.
Importantly, sketchers drew the same objects multiple times over the course of the experiment, receiving feedback about the viewer's response after each trial (Fig.~\ref{task_stimuli}B).
This repeated reference game design thus allowed us to track both changes in how well each dyad communicated, as well as changes in the content of their drawings over time.

\begin{figure}
\begin{center}
\includegraphics[width=0.5\linewidth]{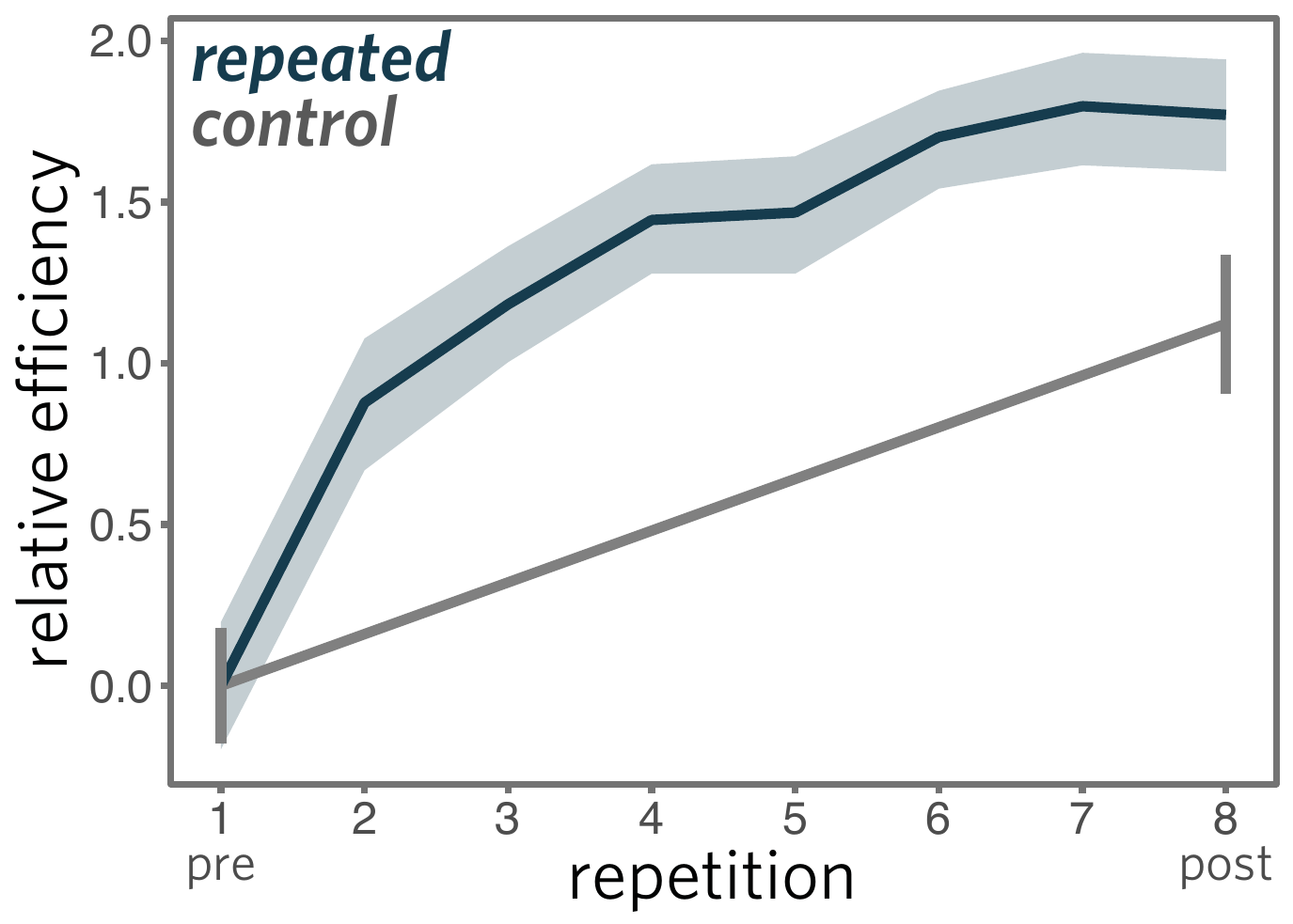}
\caption{Communication efficiency across repetitions. Efficiency combines both speed and accuracy, and is plotted relative to the first repetition. Error ribbons represent 95\% CI.}
\label{refgame_bis}
\end{center}
\end{figure}

\subsection{Improvement in communicative efficiency}

\begin{figure*}[t!]
\includegraphics[width=0.65\linewidth]{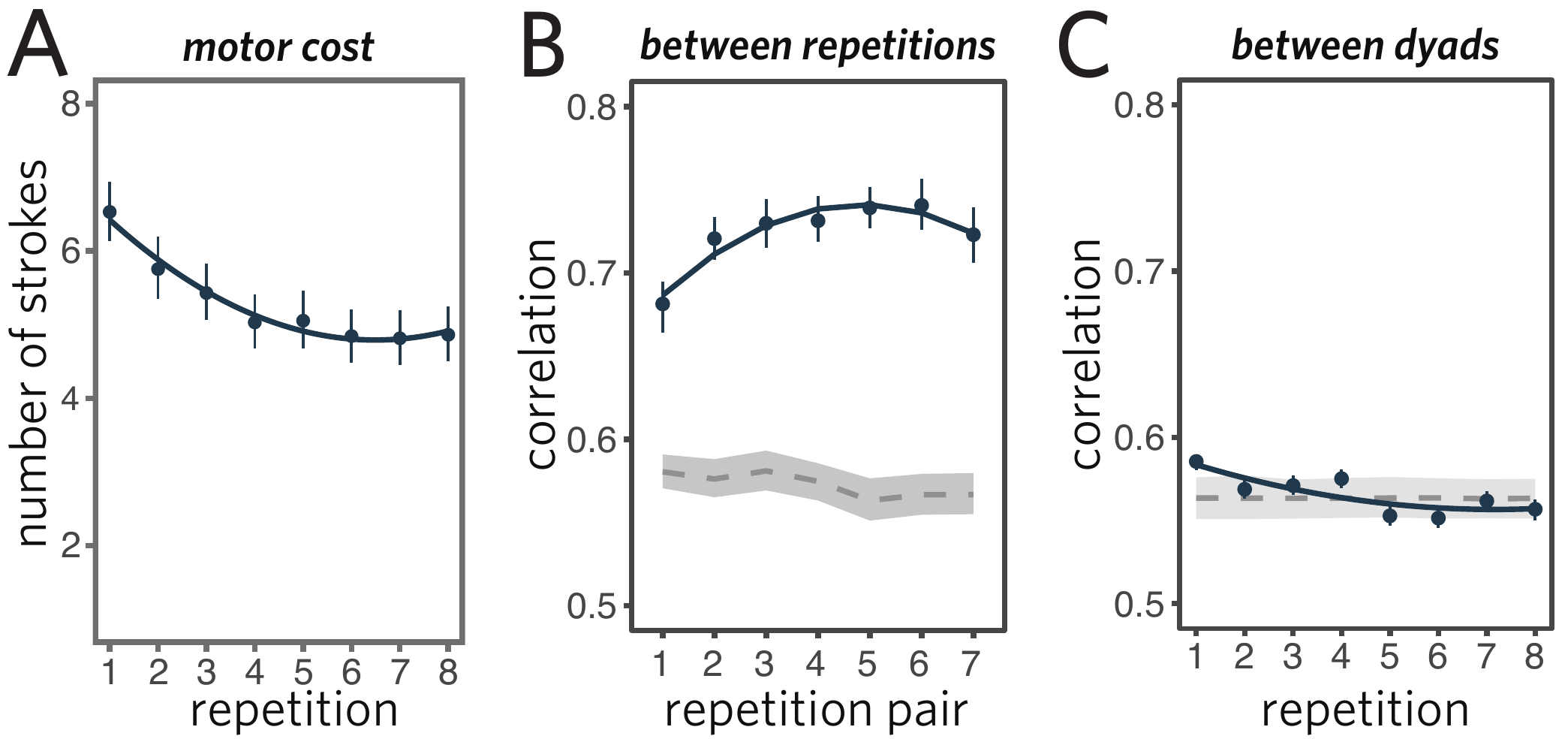}
\centering
\caption{(A) Decrease in number of strokes used to produce drawings across repetitions. (B) Increased consistency between successive drawings throughout an interaction. (C) Increased dissimilarity beween drawings of same object from different interactions. Error ribbons represent 95\% CI, dotted lines represent permuted baseline.}
\vspace{2em}
\label{within-across}
\end{figure*}

Given that the focus of our study was on changes in communication behavior over time, we sought to first verify that dyads were generally able to perform the visual communication task.
We found that even the first time sketchers drew an object, viewers correctly identified it at rates well above chance (76\%, chance = 25\%), suggesting that they were engaged with the task but not yet at ceiling performance. 
In order to measure how well dyads learned to communicate throughout the rest of their interaction, we used a measure of communicative efficiency \cite<the \emph{balanced integration score,}>{Liesefeld2018} that takes both accuracy (i.e., proportion of correct viewer responses) and response time (i.e., latency before viewer response) into account.
This efficiency score is computed by first $z$-scoring accuracy and response time for each drawing within an interaction, in order to map different interactions onto the same scale.
We then combined these measures by subtracting the standardized response time from standardized accuracy.
Efficiency is highest when dyads are both fast and accurate, and lowest when they make more errors and take longer, relative to their own performance on other trials.
We found that communicative efficiency reliably improved across repetitions of each object, $b=0.5,~t=13.5,~p<0.001$; Fig.~\ref{refgame_bis}.
Similar results were found when examining only response times ($b=-1.5,~t=-11.5,~p<0.001$) or accuracy ($b=0.46,~z=6.5,~p < 0.001$) alone, indicating that participants had achieved greater efficiency by becoming both faster and more accurate. 
One straightforward explanation for these gains is that sketchers were able to use fewer strokes per drawing to achieve the same level of viewer recognition accuracy.
Indeed, we found that the number of strokes in drawings of repeated objects decreased steadily as a function of repetition ($b = -0.216, ~t = -6.00, p <0.001$; Fig.~\ref{within-across}A). 
Overall, these results show that dyads were able to visually communicate about these objects more efficiently across repetitions.

\subsection{Improvements in communication are object-specific}

While these performance gains are consistent with the possibility that participants had developed ways of depicting each object that were dependent on previous attempts to communicate about that object, these gains may also be explained by general benefits of task practice. 
To tease apart these potential explanations, we also examined changes in communication performance for a set of control objects that were drawn only once at the beginning (\emph{pre} phase) and at the end (\emph{post} phase;  Fig.~\ref{task_stimuli}C).
In the \emph{pre} phase, there was no difference in accuracy between repeated and control objects (75.7\% repeated, 76.1\% control, mean difference: 0.3\%, bootstrapped CI: $[-7\%, 7\%]$), which was expected, as objects were randomly assigned to repeated and control conditions.
To evaluate changes in communicative efficiency, we fit a linear mixed-effects model including random intercepts, slopes, and interactions for each dyad.
We found that communicative efficiency reliably increased overall between the \textit{pre} and \textit{post} phases ($b = 0.72,~t = 14.6,~p <0.001$), suggesting at least some general benefit of task practice. 
Critically, however, we also found a reliable interaction between phase and condition: communicative efficiency improved to a greater extent for repeated objects than control objects ($b = -0.16, ~t = -3.17,~p = 0.002$; see Fig.~\ref{refgame_bis}).
Analyzing changes in raw accuracy yielded similar results (control: $+7.1\%$, repeated: $+14.5\%$; interaction: $b=-1.9,z=-2.8,p=0.005$).
Together, these data provide evidence for benefits of repeatedly communicating about an object that accrue specifically to that object.

An intriguing possibility is that dyads achieved such benefits by developing \textit{ad hoc} graphical conventions establishing what was sufficient and relevant to include in a drawing to support rapid identification of objects they repeatedly communicated about. 
To investigate this possibility, we examined how the drawings themselves changed throughout each interaction, hypothesizing that successive drawings of the same object produced within an interaction changed less over time as dyads converged on consistent ways of communicating about each object. 
For these analyses, we capitalized on recent work validating the use of image features extracted by deep convolutional neural network (DCNN) models to measure visual similarity between drawings \cite{fan2018common}.
Specifically, we used a DCNN architecture known as VGG-19 \cite{simonyan2014very} to extract feature vectors from pairs of successive drawings of the same object made within the same interaction (i.e.~repetition $k$ to $k+1$), and computed the correlation between each pair of feature vectors.
A mixed-effects model with random intercepts for both object and dyad revealed that the similarity between successive drawings increased throughout each interaction ($b = 0.53,~t = 5.03$; Fig.~\ref{within-across}B), providing support for the notion that dyads converged on increasingly consistent ways to communicate about each object. 


\subsection{Performance gains depend on shared interaction history}

One way of understanding our results so far is that the need to repeatedly refer to certain objects is sufficient to explain how the way sketchers depicted them changed over time. 
However, these objects did not appear in isolation, but rather as part of a communicative context including the viewer and the other, distractor objects. 
How did this communicative context influence the way drawings conveyed meaning about the target object across repetitions?  
To investigate this question, we conducted a follow-up \emph{recognition} experiment (see Fig.~\ref{task_stimuli}D) including two control conditions to estimate how recognizable these drawings were to naive viewers, outside the communicative context in which they were produced.
Participants in the \emph{yoked} control group were shown a sequence of drawings taken from a single interaction, closely matching the experience of viewers in the communication experiment.
Participants in the \emph{shuffled} control group were instead shown a sequence of drawings pieced together from many different interactions, thus disrupting the continuity experienced by viewers paired with a single sketcher.
Insofar as interaction-specific shared knowledge contributed to the efficiency gains observed previously, we hypothesized that the second group would not improve as much over the course of the experimental session as the first group would.
Critically, groups in both control conditions received exactly the same amount of practice recognizing drawings and performed the task under the same incentives to respond quickly and accurately.
Thus any differences in performance between these groups is attributable to the role of context in guiding the interpretability of a drawing, and in particular the accumulation of experience in the same communicative context. 


\begin{figure}[t!]
\begin{center}
\includegraphics[width=0.5 \linewidth]{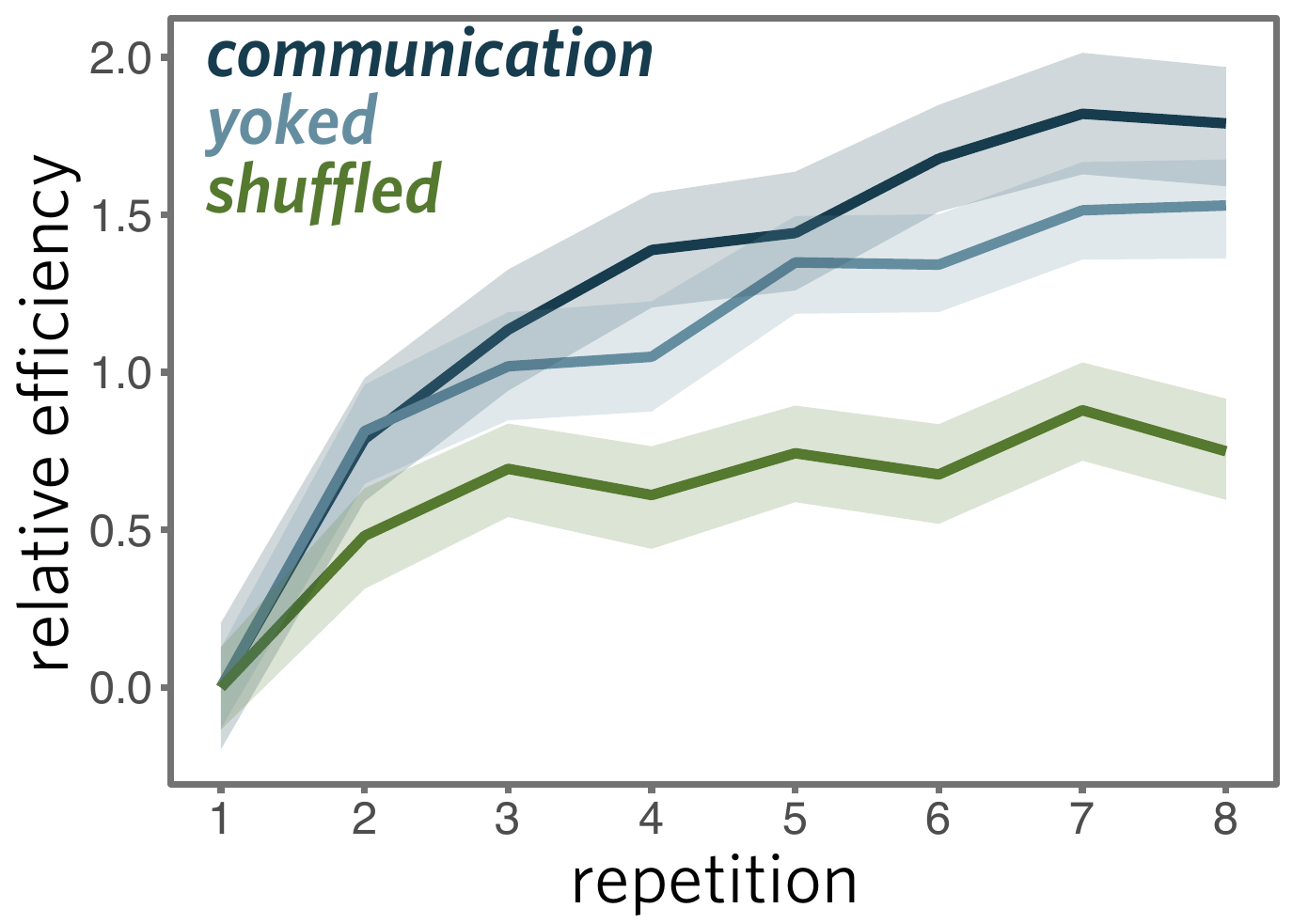}
\caption{Comparing drawing recognition performance between viewers in communication experiment with those of yoked and shuffled control groups. Error ribbons represent 95\% CI.}
\label{recog_bis}
\end{center}
\end{figure}




We compared the yoked and shuffled groups by measuring changes in recognition performance across successive repetitions using the same efficiency metric we previously used.
We estimated the magnitude of these changes by fitting a linear mixed-effects model that included group (yoked vs. shuffled), repetition number (i.e., first through eighth), and their interaction, as well as random intercepts and slopes for each participant.
While we found a significant increase in recognition performance across both groups ($b = 0.18, ~t = 12.8, ~p < 0.001$), 
we also found a large and reliable interaction:
yoked participants improved their efficiency to a substantially greater degree in than shuffled participants ($b = 0.10, ~t = 4.9, ~p<0.001$; Fig.~\ref{recog_bis}).
Examining accuracy alone yielded similar results: the yoked group improved to a greater degree across each experimental session (yoked: +15.8\%, shuffled: +5.6\%).
Taken together, these results suggest that third-party observers in the yoked condition who viewed drawings from a single interaction were able to take advantage of this continuity to more accurately identify what successive drawings represented.
While observers in the shuffled condition still improved over time, being deprived of this interaction continuity made it more difficult to interpret later drawings.

These results suggest that the graphical conventions discovered by different dyads were increasingly opaque to outside observers, consistent with prior work while additionally controlling for confounds in earlier studies, such as task practice \cite{garrod_foundations_2007}. 
Such results could arise if early drawings were more strongly constrained by the visual properties of a shared target object, but later drawings diverged as different dyads discovered different equilibria in the space of viable graphical conventions.
Under this account, drawings of the same object from different dyads would become increasingly dissimilar from each other across repetitions.
We again tested this prediction using high-level visual representations of each drawing derived from a deep neural network.
Specifically, we computed the mean pairwise similarity between drawings of the same object within each repetition index, but produced in different interactions.
In other words, we considered all interactions in which a particular object was repeatedly drawn, then computed the average similarity between drawings of that object made by different sketchers at each point in the interaction.
In a mixed-effects regression model including linear and quadratic terms, as well as random slopes and intercepts for object and dyad, we found a small but reliable negative effect of repetition on between-interaction drawing similarity ($b = -1.4, ~t = -2.5$; Fig.~\ref{within-across}C). 
We also conducted a permutation test to compare this $t$ value with what would be expected from scrambling drawings across repetitions for each sketcher and target object and found that the observed slope was highly unlikely under this distribution $(CI = [-0.57, 0.60],~p~<~0.001)$.
Taken together, these results suggest that drawings of even the same object can diverge over time when produced in different communicative contexts.






Unlike viewers in the interactive visual communication experiment, participants in the yoked condition made their decision based only on the whole drawing and were unable to interrupt or await additional information if they were still uncertain.
Sketchers could have used this feedback to modify their drawings on subsequent repetitions.
As such, comparing the yoked and original communication groups provides an estimate of the contribution of these viewer feedback channels to gains in performance \cite{schober_understanding_1989}.
In a mixed-effects model with random intercepts, slopes, and interactions for each unique trial sequence, we found a 
strong main effect of repetition ($b = 0.23, ~t = 12.8,~p < 0.001$), as well as a weaker but reliable interaction with group membership ($b = -0.05, ~t = -2.2, ~p = 0.032$, Fig.~\ref{recog_bis}), showing that the yoked group improved at a more modest rate than viewers in the original communication experiment had.
To better understand this interaction, we further examined changes in the accuracy and response time components of the efficiency score.
We found that while viewers in the communication experiment were more accurate than yoked participants overall (communication: 88\%, yoked: 75\%), 
\emph{improvements} in accuracy over the course of the experiment were similar in both groups (communication: +14.5\%, yoked: +15.8\%).
The interaction instead appeared to be driven by differential reductions in response time between the first and final repetitions (communication: 10.9s to 5.84s; yoked: 4.66s to 3.31s).
These reductions were smaller in the yoked group, given that these participants did not need to wait for each stroke to appear before making a decision, and thus may have already been closer to floor.






\subsection{Sketchers preserve visual properties that are diagnostic of object identity}






\begin{figure*}[th!]
\includegraphics[width=.8\linewidth]{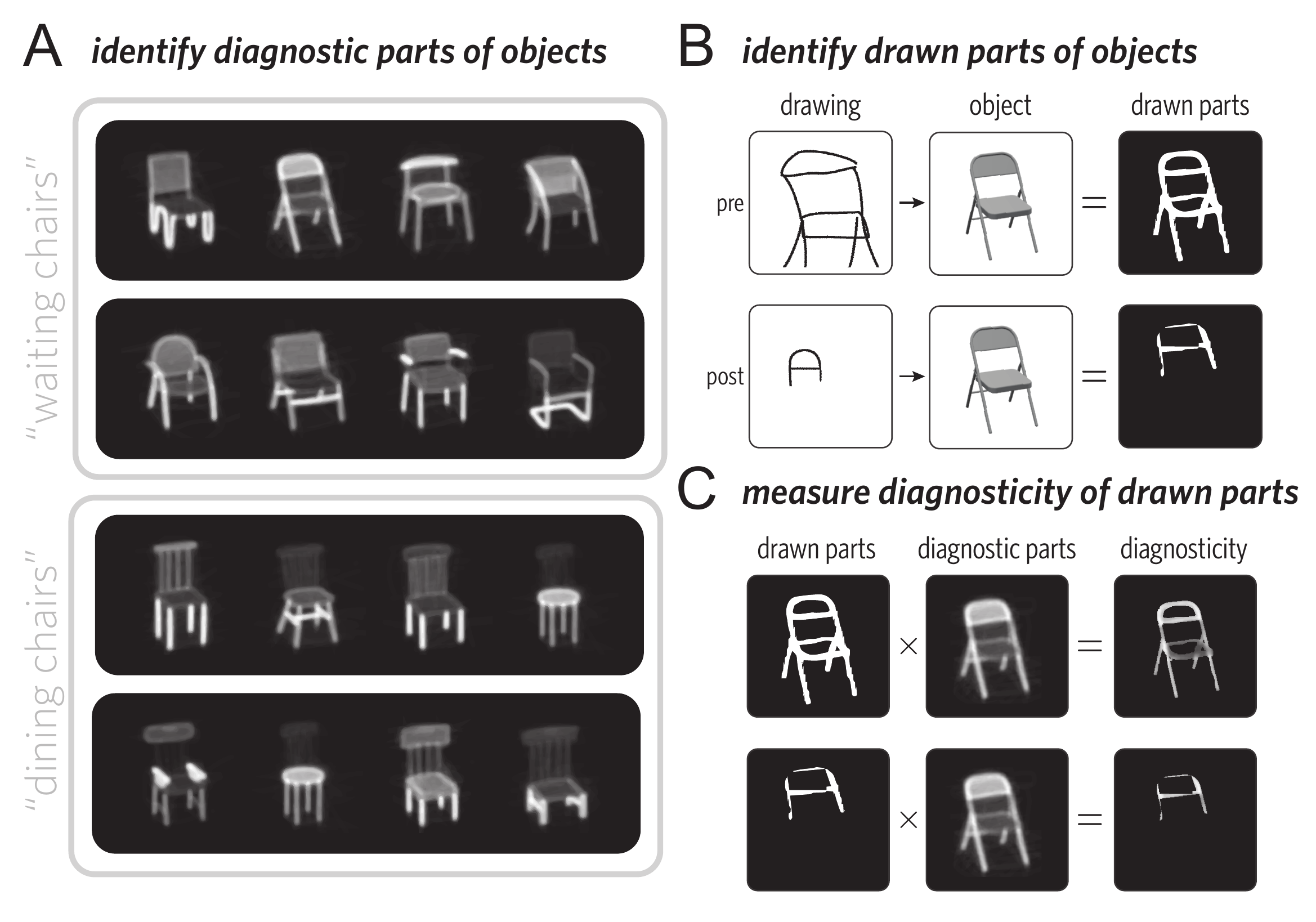}
\centering
\caption{(A) Annotators indicated which parts of an object were most diagnostic in context (brighter regions are more diagnostic), yielding a graded diagnosticity heatmap for each object. (B) A separate group of annotators also indicated which parts of objects were depicted in each drawing, yielding a binary image mask for each drawing. (C) Mean diagnosticity for a drawing was computed by averaging the diagnosticity values of all pixels in the object diagnosticity map that appeared in that drawing.}
\vspace{2em}
\label{fig:diagnosticity_methods}
\end{figure*}

Our results in the previous section suggest that viewers depend on a combination of visual information and social information to successfully recognize drawings.
Specifically, we found that it was increasingly difficult for viewers in the shuffled condition to make sense of drawings in the absence of shared interaction history with a consistent social partner.
While these findings focused primarily on the cognitive mechanisms employed by the viewer, the increasing sparsity of the drawings suggest that decisions about drawing \emph{production} may also be guided by a combination of visual and social information.
In this section we ask: Why was some visual information preserved during the formation of these graphical conventions while other information was dropped?
One possibility is that these choices are mostly arbitrary: given a sufficiently long interaction history to establish the association, any scribble could in principle be used to refer to any object.
An alternative possibility is that these choices are systematically driven by visual information: sketchers may preserve information about \emph{diagnostic} or \emph{salient} parts of the target object, rather than omitting visual information in an arbitrary fashion.
For example, in the contexts shown in Fig.~\ref{fig:diagnosticity_methods}A, the folding chair (top row, second from left) has a seat that is similar to the distractors, but a distinctive backrest and set of legs.
If sketchers are under pressure to produce informative drawings for their partner in context \cite{fan2020pragmatic, hawkins2020characterizing}, their conventions may come to reflect these pressures.

To test this hypothesis, and obtain reliable estimates of diagnosticity in context, we required a large number of drawings for a smaller set of contexts.
Instead of randomly sampling different contexts for each dyad, as before, we adapted our referenge game paradigm to only include two pre-generated contexts for every dyad, which were counter-balanced across the \emph{repeated} and \emph{control} conditions. 
We also made one important modification to our experimental design design to address a potential confound.
Rather than allowing the viewer to interrupt the sketcher with an early response, we required the sketcher to click a ``done'' button when they were ready to show their drawing to the viewer.
Here, drawing duration is purely a function of the sketcher's independent decisions about what needs to be included in a drawing, whereas in our original design, it was a joint combination of the sketcher's decision and the viewer's decision threshold for when to interrupt.
That is, it was possible in the original design that any apparent effects of conventionalization were purely driven by the viewer, with the sketcher simply following a heuristic to continue adding more detail until the viewer made a decision.
Aside from these changes, the design was identical to the original repeated reference game. 

We recruited a sample of 65 additional dyads (130 participants) for this task.
In addition to providing sufficient power for our diagnosticity analyses, this new sample also provided an opportunity to conduct an internal replication to evaluate the robustness of our results (see Appendix for successful replications of our earlier analyses on these new data).
Next, we recruited a separate sample of naive annotators to determine the diagnosticity of these drawings over time. 
One group of annotators indicated which parts of objects were depicted in each drawing by painting over the corresponding regions of the target object (Fig.~\ref{fig:diagnosticity_methods}B), yielding a binary mask for each drawing.
A second group of annotators indicated which parts of objects were most diagnostic in context by painting over regions of each target object that distinguished it the most from each distractor object, yielding a graded heat map of diagnostic regions over each object (Fig.~\ref{fig:pairwise}). 

\begin{figure*}[t!]
\includegraphics[width=.4\linewidth]{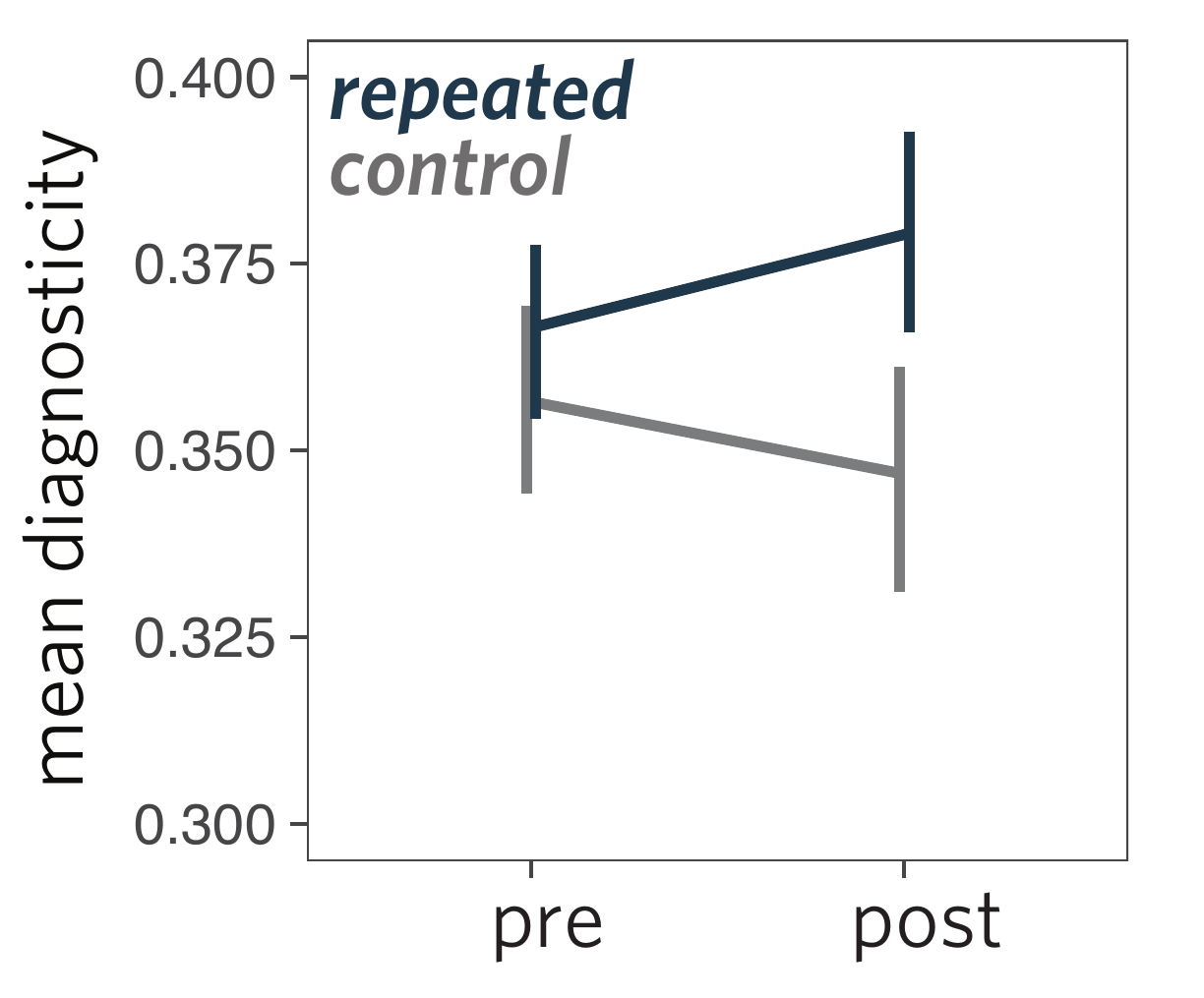}
\centering
\caption{Changes in mean diagnosticity of drawn parts over time. Error bars represent bootstrapped 95\% confidence intervals.}
\vspace{2em}
\label{fig:diagnosticity_results}
\end{figure*}

To measure changes in the diagnosticity of drawings over time, we took the intersection of these annotation maps for each drawing (see Fig.~\ref{fig:diagnosticity_methods}C).
We then took the average diagnosticity value per pixel in the combined stroke map to control for the overall size of the drawing, a metric reflecting how much the sketcher had selectively prioritized diagnostic parts of the object overall.
Our primary hypothesis concerned differential changes in diagnosticity over time. 
Insofar as new graphical conventions are shaped by communicative context, gradually depicting the most distinctive regions of the image while omitting less distinctive regions, we predicted that the repeated drawings would \emph{increase} in diagnosticity between the pre- and post- phases. 
Meanwhile, to the extent that these changes in diagnosticity depend on having communicated repeatedly about an object, we predicted that the diagnosticity of control drawings would remain stable over time.
To test these hypotheses, we conducted a mixed-effects regression analysis on diagnosticity values for each drawing.
We included fixed effects of phase (pre vs. post) and condition (repeated vs. control) as well as their interaction.
While the maximal random effects structure did not converge, we were able to include intercepts and main effects for each sketcher and each target object. 
Consistent with our hypothesis, we found a significant interaction ($b=-0.05,~t=-3.4,~p < 0.001$, Fig.~\ref{fig:diagnosticity_results}): objects in the repeated condition became increasingly diagnostic as they became sparser, relative to those in the control condition. 

\section{Discussion}

The puzzle of pictorial meaning has long resisted reductive explanations. 
Classical theories have either argued that a picture's meaning is primarily determined by visually resembling entities in the world, or by appealing to socially mediated conventions.
However, these theories fail to explain the full range of pictures that people produce.
In this paper, we proposed an integrative cognitive theory where both resemblance and conventional information jointly guide inferences about what pictures mean.
We evaluated this theory using a Pictionary-style communication game in which pairs of participants developed novel graphical conventions to depict objects more efficiently over time. 
Our theory predicted that viewers would initially rely on visual resemblance between the drawing and images to successfully determine the intended referent, but rely increasingly on experience from earlier communicative exchanges even as direct resemblance decreased.
We tested these predictions by manipulating the amount and type of socially mediated experience available to the viewer: we varied how often each object had been drawn throughout an interaction and whether the drawings were produced by the same individual.
We found that viewers improved to a greater degree for objects that had been drawn more frequently; conversely, viewers had greater difficulty recognizing sequences of drawings produced by different individuals.
We further tested the prediction that sketchers in our task would also increasingly rely on shared experience with a specific viewer, and found that people produced progressively simpler drawings that prioritized the most diagnostic visual information about the target object's identity.
Taken together, our findings suggest that visual resemblance forms a foundation for pictorial meaning, but that shared experiences promote the emergence of depictions whose meanings are increasingly determined by interaction history rather their visual properties alone.

\begin{figure*}[t!]
\includegraphics[width=.8\linewidth]{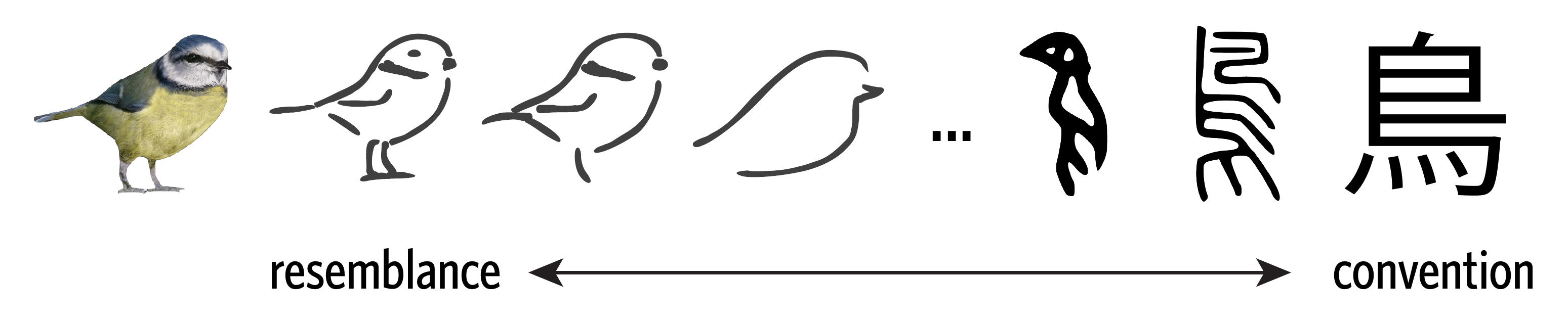}
\centering
\caption{Our findings support the notion that both visual resemblance and socially mediated conventions jointly guide inferences about pictorial meaning.}
\vspace{2em}
\label{fig:discussion_schematic}
\end{figure*}


There are several important limitations of the current work that future studies should address to further evaluate this integrative theory of pictorial meaning (see Fig.~\ref{fig:discussion_schematic}).
First, here we focused on how people use drawings to communicate about the identity of a visual object.
As such, we were able to leverage existing techniques for encoding high-level visual features of both drawings and objects into the same latent feature space to operationalize their visual resemblance \cite{fan2018common}. 
However, people also produce pictures to communicate about non-visual concepts, such as semantic associations \cite{garrod_foundations_2007,schloss2018color}, number \cite{chrisomalis2010numerical,holt2021improvised}, and causal mechanisms \cite{bobek2016creating,huey2021semantic}.
It is unclear whether the same general-purpose visual processing mechanisms will be sufficient to explain how graphical conventions emerge to convey these more abstract concepts. 
To the extent that general-purpose visual encoding models can easily generalize to a particular `non-visual' concept without relying upon \emph{ad hoc} associative learning, then visual resemblance may play a stronger role in explaining how that abstract concept is grounded in graphical representations of them. 
On the other hand, if and when such associative learning mechanisms are necessary above and beyond such generic visual processing mechanisms to explain the mapping between a picture and an abstract concept (e.g., ``42'' or $\rightarrow$), then conventionality may play a stronger role for explaining how such pictures become meaningful in context, consistent with existing descriptive accounts of what distinguishes symbols from icons \cite{gelb1963study,wescott1971linguistic,verhoef2016iconicity,perlman2015iconicity,peirce1974collected}.
There is thus substantial mechanistic clarity to be gained by developing more robust computational models that can operate on a broader range of images to predict a greater variety of abstract meanings beyond the identity of individual objects.



A second important direction for future work would be to explore why drawings are produced at different points along the resemblance-convention continuum at all. 
In other words, if resemblance is sufficient, why rely upon socially mediated experience at all?
Our paradigm suggests that production cost may be one important factor that driving such behavior.
Recent computational models of visual communication have found that how costly a drawing is to produce (i.e., in time/ink) is critical for explaining the way people spontaneously adjust the level of detail to include in their drawings in one-shot visual communication tasks \cite{fan2020pragmatic}.
We expect that the consequences of this intrinsic preference for less costly drawings may be compounded across repetitions, as the accumulation of feedback and interaction history allows people to continue to be informative with fewer strokes, effectively increasing the capacity of the communication channel \cite{HawkinsFrankGoodman17_ConventionFormation}.
The magnitude of such implicit production costs may vary across individuals, however, motivating our use of explicit incentives for all participants to complete trials efficiently.
Further work should explore other considerations driving the tradeoffs between relying on resemblance-based and convention-based cues, including the reliability of resemblance-based information, the complexity of the target concept, and the availability of social feedback.
Finally, our framework for pictorial meaning may help illuminate why visual communication has been such a uniquely powerful vehicle for the cultural transmission of knowledge across so many cultures.
In particular, our work suggests that the ability to easily rely on resemblance-based cues to meaning gives the visual modality unique advantages over other modalities for conveying certain information.
In other words, the cognitive mechanisms supporting successful visual communication may be rooted in our shared visual systems, facilitating communication between members of different language communities, even in the absence of shared graphical conventions.
Advancing our knowledge of the cognitive mechanisms underlying pictorial meaning may thus lead to a deeper understanding of how humans are capable of seamlessly integrating such a huge variety of graphical and symbolic representations to think and communicate. 

\section{Methods}

\subsection{Reference game experiment}

\paragraph{Participants} We recruited 138 participants from Amazon Mechanical Turk, who were paired up to form 69 dyads to play a drawing-based reference game \cite{Hawkins15_RealTimeWebExperiments}.
For our diagnosticity analyses, which required higher power for a smaller number of specific contexts, we recruited an additional 130 participants (65 dyads).
Participants were provided a base compensation of \$1.50 for participation and were able to earn an additional \$1.60 in bonus pay based on task performance.
In this and subsequent experiments, participants provided informed consent in accordance with the Stanford IRB.

\paragraph{Stimuli}

In order to make our task sufficiently challenging, we sought to construct visual contexts consisting of objects whose members were both geometrically complex and visually similar.
To accomplish this, we sampled objects from the ShapeNet \cite{chang2015shapenet}, a database containing a large number of 3D mesh models of real-world objects. 
We restricted our search to 3096 objects belonging to the \texttt{chair} class, which is among the most diverse and abundant in ShapeNet.
To identify groups of visually similar objects, we employ neural-network based encoding models to extract high-level feature representations of images. 
Specifically, we used the PyTorch implementation of the VGG-19 architecture pre-trained to perform image classification on the ImageNet database \cite{simonyan2014very,deng2009imagenet,paszke2017automatic}, an approach that has been validated in prior work to provide a reasonable proxy for human perceptual similarity ratings between images of objects \cite{peterson2018evaluating,kubilius2016deep}.
This feature extraction procedure yields a 4096-dimensional feature vector for each rendering, reflecting activations in the second fully-connected layer (i.e., \texttt{fc6}) of VGG-19, a higher layer in the network. 
We then applied dimensionality reduction (PCA) and $k$-means clustering on these feature vectors, yielding 70 clusters containing between 2 and 80 objects each.
Among clusters that contained at least eight objects, we manually identified two visual categories containing eight objects each, which roughly correspond to `dining chairs' and `waiting-room chairs.'

\paragraph{Design}
For each dyad, two sets of four objects were randomly sampled to serve as communication contexts: one was designated the \emph{repeated} set while the other served as the \emph{control} set\footnote{In half of the dyads, the four control objects were from the same stimulus cluster as repeated objects; in the other half, they were from different clusters. The rationale for this was to support investigation of between-cluster generalization in future analyses. In current analyses, we collapse across these groups.}.
Our second sample simply restricted the stimuli to two fixed sets of four objects, which were counter-balanced to \emph{repeated} and \emph{control}, instead of randomly sampling sets, in order to obtain sufficient observations per set.
The experiment consisted of three phases.
During the \textit{repetition} phase, there were six repetition blocks of four trials, and each of the four repeated objects appeared as the target once in each repetition block.
In a \textit{pre} phase at the beginning of the experiment and a \textit{post} phase at the end, both repeated and control objects appeared once as targets (in their respective contexts) in a randomly interleaved order.

\paragraph{Task Procedure}

Upon entering the session, one participant was assigned the sketcher role and the other was assigned the viewer role. 
These role assignments remained the same throughout the experiment.
On each trial, both participants were shown the same set of four objects in randomized locations.
One of the four objects was highlighted on the sketcher's screen to designate it as the target.
Sketchers drew using their mouse cursor in black ink on a digital canvas embedded in their web browser ($300 \times 300$ pixels; pen width = 5px).
Each stroke was rendered on the viewer's screen in real time and sketchers could not delete previous strokes.
The viewer aimed to select the true target object from the context of four objects as soon as they were confident of its identity, and both participants received immediate feedback: the sketcher learned when and which object the viewer had clicked, and the viewer learned the true identity of the target.
Participants were incentivized to perform both quickly and accurately.
They both earned an accuracy bonus for each correct response, and the sketcher was required to complete their drawings in 30 seconds or less.
If the viewer responded correctly within this time limit, participants also received a speed bonus inversely proportional to the time taken until the response.
There was only one procedural difference in our second, replication sample: instead of allowing the viewer to interrupt the production of the drawing at any point (as in Pictionary), we required them to wait until the sketcher decided to finish and press a ``Done'' button.
This change removed potential confounds between the speaker's decision-making and the listener's decision-making, as the drawing time is now purely under the speaker's control.

\subsection{Recognition experiments}

\paragraph{Participants}

We recruited 245 participants via Amazon Mechanical Turk and excluded data from 22 participants who did not meet our inclusion criterion for accurate and consistent response on attention-check trials, leaving a sample of 223 participants (106 in yoked, 117 in shuffled).
For our internal replication, conducted on the secondary dataset collected for our diagnosticity analyses, we obtained data from an additional 225 participants, after exclusions (100 in yoked, 125 in shuffled). 

\paragraph{Design \& Procedure}

On each trial, participants were presented with a drawing and the same set of four objects that accompanied that drawing in the original visual communication experiment.
They also received the same accuracy and speed bonuses as viewers in the communication experiment.
To ensure task engagement, we included five identical attention-check trials that appeared once every eight trials.
Each attention-check trial presented the same set of objects and drawing, which we identified during piloting as the most consistently and accurately recognized by naive participants.
Only participants who responded correctly on at least four out of five of these trials were retained in subsequent analyses.
Each participant was randomly assigned to one of two conditions: a \textit{yoked} group and a \textit{shuffled} group.
Each yoked participant was matched with a single interaction from the original cohort and viewed 40 drawings in the same sequence the original viewer had.
Those in the shuffled group were matched with a random sample of 10 distinct interactions from the original cohort and viewed four drawings from each in turn, which appeared within the same repetition block as they had originally.
For example, if a drawing was produced in the fifth repetition block in the original experiment, then it also appeared in the fifth block for shuffled participants.

\subsection{Model-based analyses of drawing features}

To extract high-level visual features of drawings, we used the same PyTorch implementation of the VGG-19 architecture that we used to cluster our stimuli. 
Using these learned feature representations to approximate human judgments about the high-level visual properties of drawings has been validated in prior work \cite{fan2018common}.
This feature extraction procedure yields a 4096-dimensional vector representation for drawings of every object, in every repetition, from every interaction.
Using this feature basis, we compute the similarity between any two drawings as the Pearson correlation between their feature vectors (i.e., $s_{ij} =  \nicefrac{cov(\vec{r}_{i}, \vec{r}_{j})}{\sqrt{var(\vec{r}_{i}) \cdot var(\vec{r}_{j})}}$).

\subsection{Empirical measurement of drawing-object correspondences}

A major challenge that arises when comparing multiple drawings is the \emph{alignment problem.} 
Different drawings of the same object may be made at different scales, or translated with some spatial offset on the canvas.
Additionally, when different drawings depict different partial views of an object, it is not straightforward to determine how exactly strokes in one drawing should map onto strokes in the other.
To address these challenges, we designed a \emph{sketch-mapping} task that allows all drawings in our dataset to be projected into a common space (see Fig.~\ref{fig:object_mapping_screenshot}A).
This task was implemented with a simple annotation interface.
On one side of the screen, participants were shown a line drawing.
On the other side of the screen, they were shown a paint canvas containing the target object the drawing was intended to depict.
For each stroke in the line drawing, participants were asked to paint over the corresponding region of the target object.
We highlighted one stroke at a time, using a bright green color to visually distinguish it, and participants clicked ``Done" when they were finished making their annotation for that stroke.
Participants were not allowed to proceed to the next stroke until some paint was placed on the canvas.
To provide context, we also showed participants the history of the interaction in which the drawing appeared, so it would be clear, for instance, that an isolated half-circle corresponds to the top of the back rest, given more exhaustive earlier drawings.
They continued through all strokes of the given drawing in this way, and then proceeded to the next drawing, annotating a total of 10 different drawings in a session.
We recruited 443 participants from Amazon Mechanical Turk to perform the annotation task.
We excluded participants who consistently provided low-quality annotations (i.e. participants who made random marks on the canvas to finish the task as quickly as possible) through a combination of manual examination and response latencies.
We continued to recruit until all 2600 drawings in our dataset had at least one high-quality drawing-object correspondence map.
Finally, to reduce noise from annotators who drew outside the bounds of the image (where diagnosticity was low by definition), we applied a simple masking step in post-processing.
Specifically, we extracted a segmentation map from the ground truth image of the object to zero out any pixels in the map that corresponded to the background rather than the object.

\subsection{Empirical measurement of object-diagnostic features}

We recruited 117 participants from Amazon Mechanical Turk to provide diagnosticity maps for each target object, relative to its context.
The task interface was similar to the one we used to elicit drawing-object correspondences  (see Fig.~\ref{fig:object_mapping_screenshot}B).
A target object was displayed on the left side of the screen and a foil was displayed on the right side. 
Participants were instructed to paint over the parts of the target object that were most distinctive and different from the foil.
We elicited pairwise comparisons instead of showing the full context to reduce confusion about what was meant by ``most different'' (i.e. in a large enough context, every part of an object has some difference from at least one distractor). 
Each participant provided exactly one response for all 16 target objects used in our fixed-context experiment, and we randomly assigned participants to one of 24 possible permutations of distractors, such that different participants saw each target object paired with different distractors. 
This yielded at least 30 ratings for each pair of objects. 
To create our final heat maps (as shown in Fig.~\ref{fig:diagnosticity_methods}A), we aggregated diagnosticity ratings across the three possible foils in post-processing by taking the mean pixel intensity for each pixel.
Thus, the highest diagnosticity pixels for an object are those which were marked most consistently as distinguishing it from the most distractors. 

\section*{Data and code availability} 
All data and code for results presented in this article is available in the following GitHub repository: \url{https://github.com/cogtoolslab/graphical_conventions}. 

\section*{\bf Acknowledgments}
\small
Thanks to Mike Frank and Hyo Gweon for helpful discussion.
RDH was supported by the E. K. Potter Stanford Graduate Fellowship and the NSF Graduate Research Fellowship (DGE-114747).
MS was supported by the Masason Foundation Scholarship and the Center for the Study of Language and Information at Stanford.
JEF is supported by NSF CAREER Award \#2047191 and an ONR Science of Autonomy Award.
A subset of these findings was published as part of the Proceedings of the 41st Annual Meeting of the Cognitive Science Society in 2019.

\section*{Author contributions statement}

R.D.H., M.S., and J.E.F. designed the study, performed the experiments, and conducted analyses. R.D.H., M.S., N.D.G, and J.E.F. interpreted results and wrote the paper. 

\section*{Conflicts of Interest}

The authors declare no competing financial interests.

\bibliographystyle{apacite}

\setlength{\bibleftmargin}{.125in}
\setlength{\bibindent}{-\bibleftmargin}

\bibliography{references}

\renewcommand{\thefigure}{S\arabic{figure}}
\renewcommand{\thetable}{S\arabic{table}}
\setcounter{table}{0}
\setcounter{figure}{0}

\section*{Appendix: Results from internal replication}

Our diagnosticity analyses (Section 2.4) required a larger sample size for each context, motivating a full replication of our study.
In addition to providing data that is uniquely suited for measuring diagnosticity, this replication also provided an opportunity to internally validate our results from earlier sections in an independent sample. 
In this section, we report our findings using the same analysis pipeline on these new data ($N=65$ dyads). 
Unless otherwise stated, we used exactly the same mixed-effects model structure on both datasets.

\subsection*{2.1A: Improvement in communicative efficiency}

We computed the balanced integration score (BIS) and found a significant improvement in communicative efficiency in the repeated condition, $b=0.47, t=12.5, p<0.001$, similar to our original effect ($b=0.51, t =13.5$).
We also replicated individual effects for pure drawing time, $b=-1.7,t=-9.5, p<0.001$, and accuracy, $b=0.28,z=4.3, p<0.001$ (compared to the original effects of $b=-1.5, t = -11.5$ and $b=0.46, z=6.5$, respectively).
Finally, we replicated our finding that the number of strokes decreased, $b=-0.22, t=-4.7, p < 0.001$ (compared to the original effect of $b=-0.22, t=-6$).
Because a modification in the design prevented listeners from interrupting in our replication, this result represents a purer measure of how long the sketcher \emph{decided} to keep drawing, implying that these gains in efficiency were not solely driven by the listener's interruptions.

\subsection*{2.2A: Improvements in communication are object-specific}

Next, we included the \emph{control} condition in our analyses, replicating both the main effect of improvement between the pre-test and post-test, $b = 0.68,~t=11.6,~p < 0.001$, as well as the interaction, $b=-0.16,~t = -3.7,~p < 0.001$ (compared to the original effects of $b = 0.72,~t = 14.6$ and $b= -0.16,~t = -3.17$, respectively). 
When examining raw accuracy as our dependent variable rather than our composite BIS measure, the full mixed-effects logistic regression structure we used in the main text did not converge, so we removed the random effect of \emph{phase} and only fit dyad-level random intercepts and effects for \emph{condition}\footnote{We were able to fit the full random effect structure, including a random interaction, using the Bayesian mixed-effects regression implemented in \texttt{brms}, which yielded a similar interaction coefficient estimate, $b = -0.81$, with a 95\% credible interval of $[-1.60, -0.07]$.}.
We found a significant interaction (control: +5.8\%, repeated: +12.7\%, $b=-0.70, z = -2.0, p = 0.047$), which is a numerically large effect size but statistically weaker than our original effect (control: +7.1\%, repeated: +14.5\%, $b = -1.9, z = -2.8$). 
Finally, we again extracted high-dimensional visual features from a CNN to analyze the stability of drawings over time. 
We found a significant increase over time in the similarity of drawings made by a given sketcher on successive trials, $b = 0.57, t=5.4, p < 0.001$, consistent with our original findings ($b = 0.53, t = 5.03$). 

\subsection*{2.3A: Performance gains depend on shared interaction history}

We also conducted a replication of our control experiment using the new drawings we collected in our replication of the reference game. 
For this control experiment, we recruited 100 naive viewers for the `yoked' condition and 125 naive viewers for the `shuffled' condition.
As before, we found a significant effect of repetition on recognition performance across both conditions, $b=0.21, t=14.7, p < 0.001$, as well as a significant interaction, $b = 0.09, t = 4.9, p < 0.001$ (compared to our original effects of $b=0.18, t = 12.8$ and $b=0.10, t = 4.9$, respectively). 
Accuracy alone showed similar patterns (yoked: +15\%, shuffled: +6.6\%, compared with our original effects of 15.8\% and 5.6\%).
Finally, we examined the extent to which drawings diverge across interactions by analyzing high-dimensional visual features.
We find a significant decrease over time in the similarity of drawings produced in different interactions, $b = -2.0, t = -4.99, p = 0.001$ (consistent with our original result, $b = -1.4, t = -2.5$).

\newpage

\begin{figure*}
\includegraphics[width=.8\linewidth]{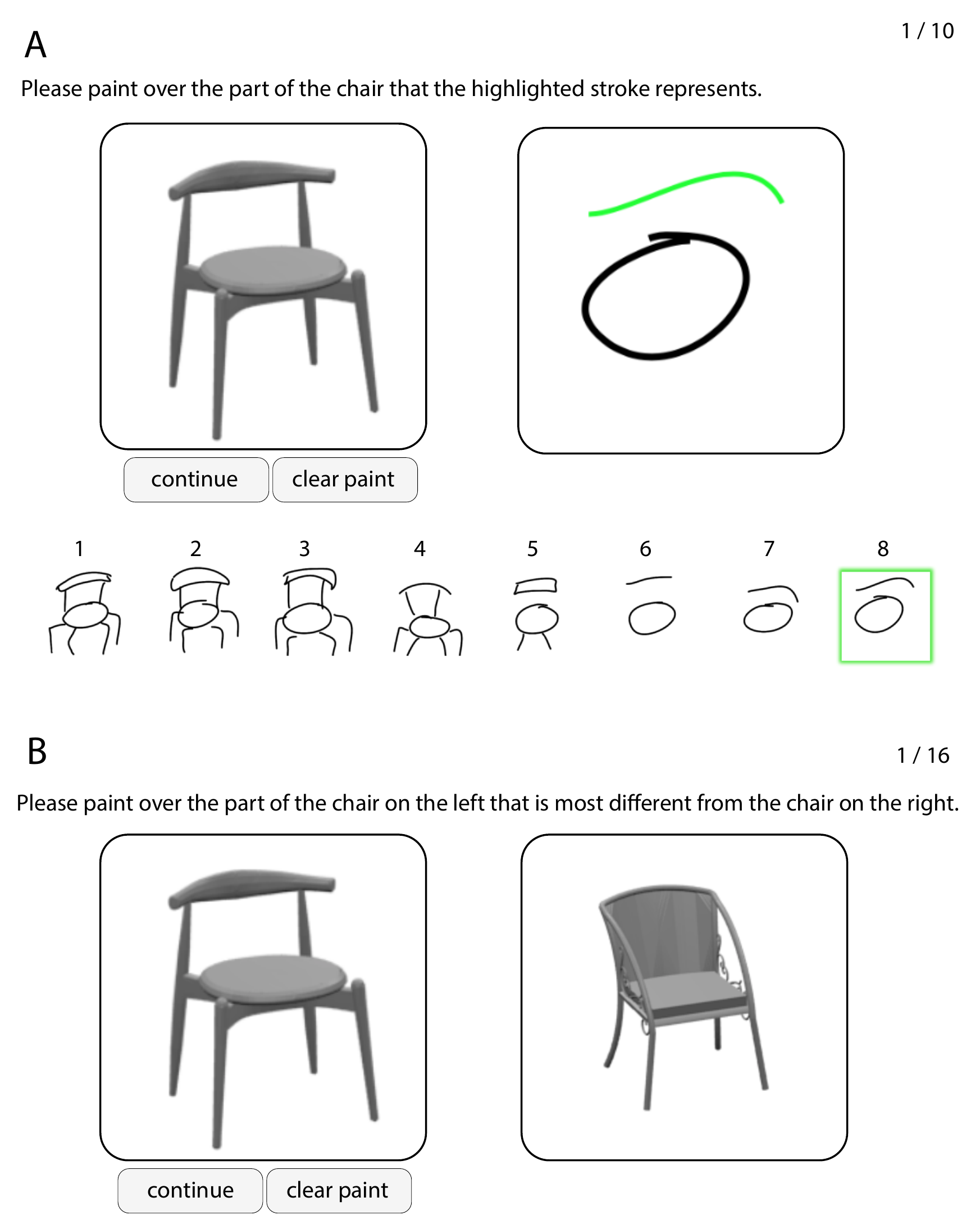}
\centering
\caption{(A) Task interface provided to annotators who indicated which parts of the object each stroke of each drawing corresponded to (B) Task interface provided to annotators who indicated which part of a target object (left) was most different from the distractor object (right). These annotations were obtained for all pairs of objects from each context, which were then aggregated to produce a graded diagnosticity map for each object.}
\label{fig:object_mapping_screenshot}
\end{figure*}

\begin{figure*}
\includegraphics[width=.8\linewidth]{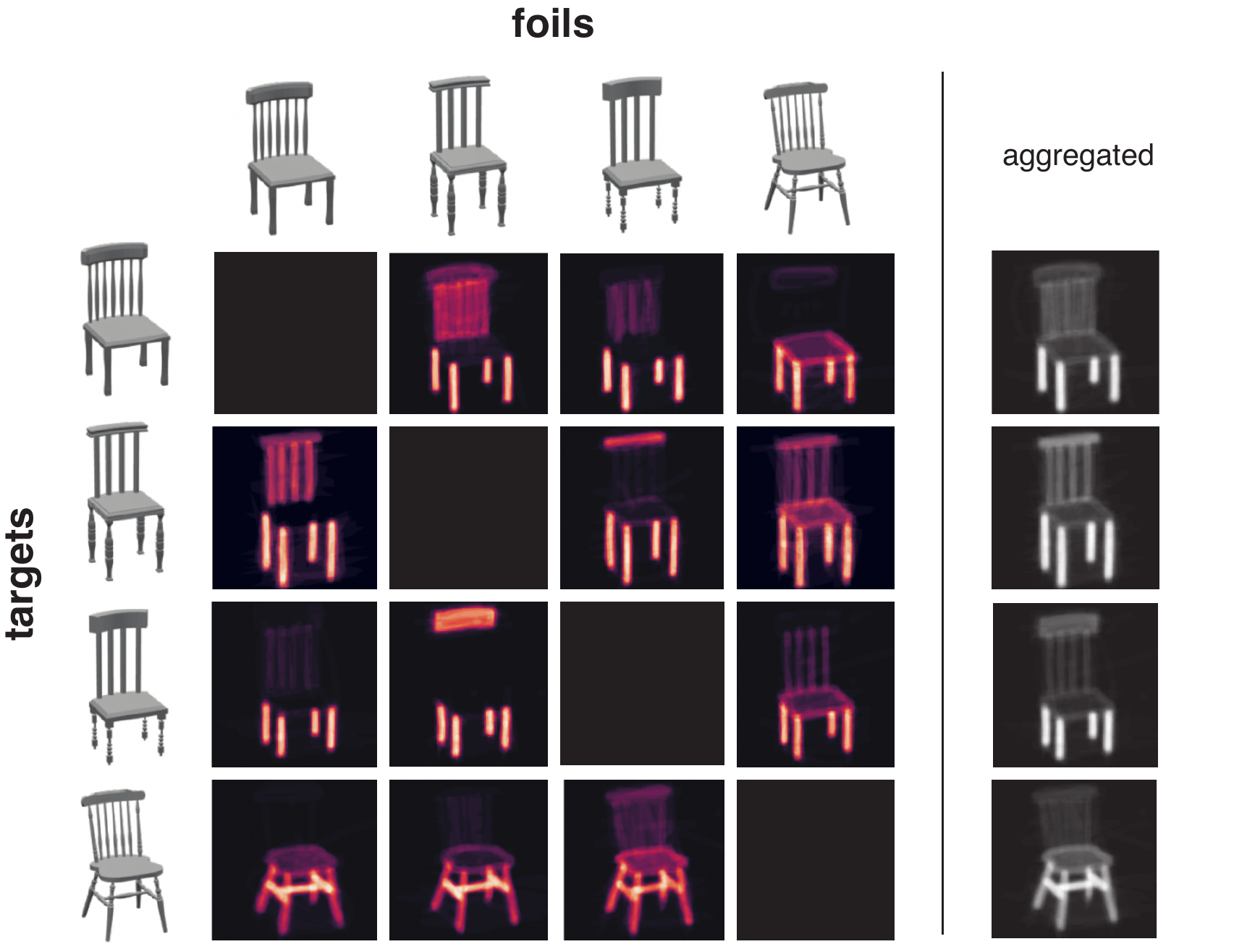}
\centering
\caption{Aggregate diagnosticity maps for each target object (rows) were constructed by combining the raw diagnosticity maps (columns) obtained from pairing the target object with each of the three distractor objects. Different regions of the target object were diagnostic for each distractor; the aggregated map captures those regions which were identified by annotators, on average, across all distractors.}
\label{fig:pairwise}
\end{figure*}
\begin{figure*}[t!]
\includegraphics[width=.2\linewidth]{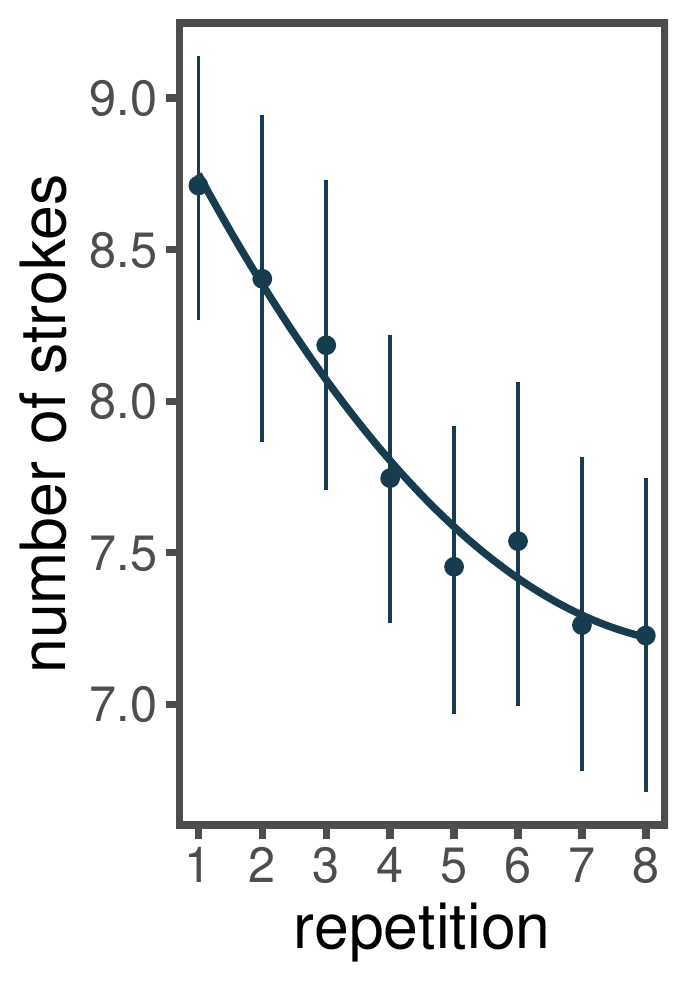}
\includegraphics[width=.4\linewidth]{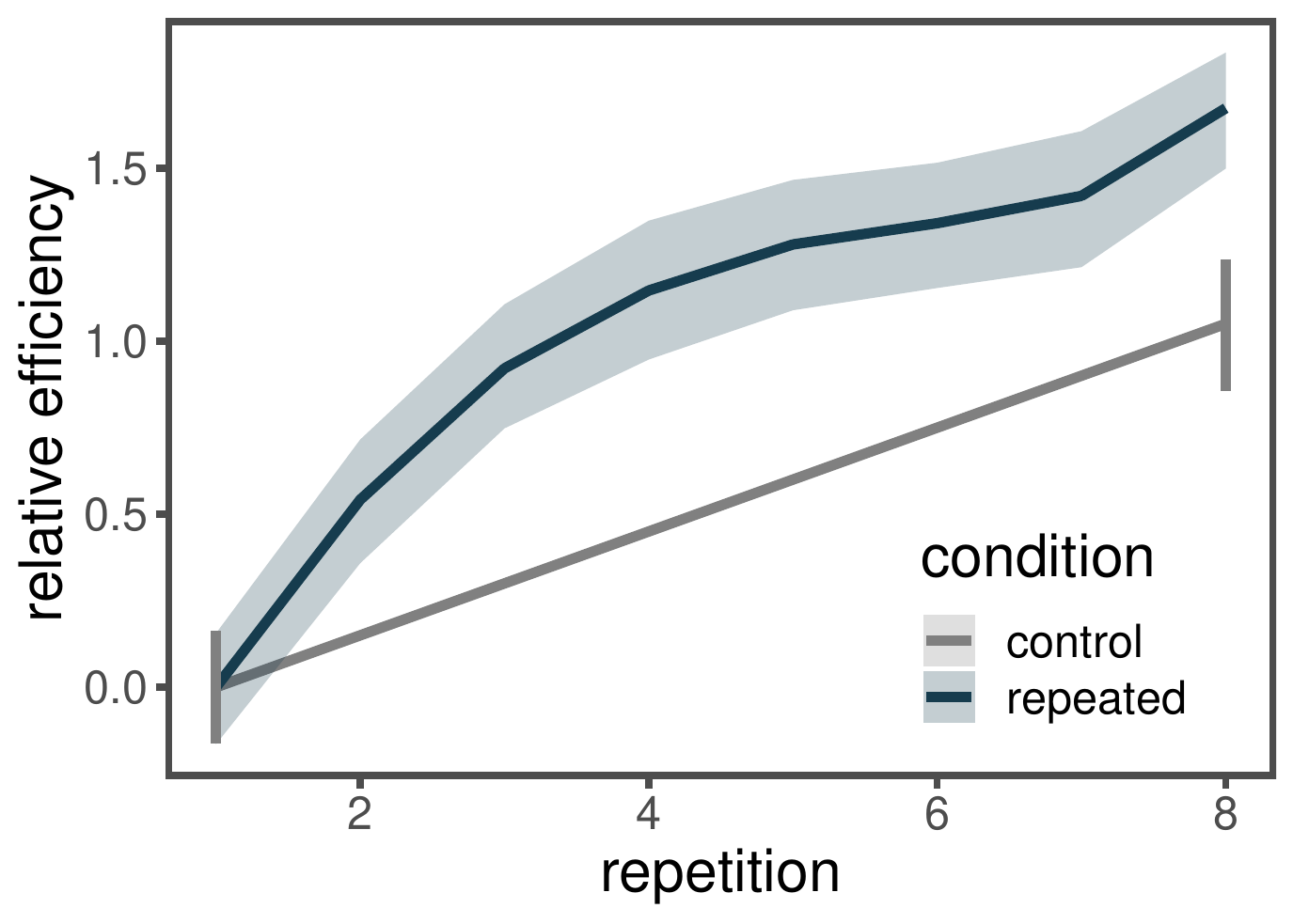}
\centering
\caption{Selected results from internal replication. \emph{Left}: The number of strokes used to produce drawings across repetitions. \emph{Right:} Communication efficiency increases across repetitions. Efficiency combines both speed and accuracy, and is plotted relative to the first repetition. Error ribbons represent 95\% CI.}
\vspace{1em}
\label{fig:object_mapping_screenshot}
\end{figure*}

\end{document}